# CRNN: A Joint Neural Network for Redundancy Detection

Xinyu Fu*, Eugene Ch'ng*, Uwe Aickelin*, Simon See†
* *The University of Nottingham Ningbo China*
*Email: {Xinyu.FU, eugene.chng, Uwe.Aickelin}@nottingham.edu.cn*
†*Nvidia Technology Center and*
*Solution Architect and Engineering*
*Asia Pacfic and Japan, Nvidia*
*Email: ssee@nvidia.com*

*Abstract*—This paper proposes a novel framework for detecting redundancy in supervised sentence categorisation. Unlike traditional singleton neural network, our model incorporates character-aware convolutional neural network (Char-CNN) with character-aware recurrent neural network (Char-RNN) to form a convolutional recurrent neural network (CRNN). Our model benefits from Char-CNN in that only salient features are selected and fed into the integrated Char-RNN. Char-RNN effectively learns long sequence semantics via sophisticated update mechanism. We compare our framework against the state-of-the-art text classification algorithms on four popular benchmarking corpus. For instance, our model achieves competing precision rate, recall ratio, and F1 score on the Google-news data-set. For twenty-news-groups data stream, our algorithm obtains the optimum on precision rate, recall ratio, and F1 score. For Brown Corpus, our framework obtains the best F1 score and almost equivalent precision rate and recall ratio over the top competitor. For the question classification collection, CRNN produces the optimal recall rate and F1 score and comparable precision rate. We also analyse three different RNN hidden recurrent cells' impact on performance and their runtime efficiency. We observe that MGU achieves the optimal runtime and comparable performance against GRU and LSTM. For TFIDF based algorithms, we experiment with word2vec, GloVe, and sent2vec embeddings and report their performance differences.

## 1. Introduction

Redundancy detection is the process pairing substrings of text to preclassified labels. With the growth of online data acquisition techniques and infrastructure, users have strong willingness to develop knowledge of the changing world. Research on redundancy detection began by traditional bag-of-words (BOW), TFIDF frequency matrix, and n-gram language modelling [1], [2]. The recent advances on the deep neural networks further push the community in the area of text classification. In accordance with Krizhevsky et al. [3] and Collobert et al. [4], CNN has shown state-of-the-art performance on computer vision and varied natural language processing (NLP) problems. Whilst traditional methods are good at encoding term association relationships (stock and market), neural models tend to preserve word similarity relationships (stock and security) [5].

We propose the utilisation of Char-CNN to sift the most salient features because of the prevalence on deep neural network and its applications to various domains like the challenging online text classification and clustering [6], [7], [8]. Char-CNN also adds confidence to accuracy by learning subword information, misspelling or grammatical errors. These issues are likely to happen in a human derived works such as newswire streams and social media data sources [6], [7], [8]. By feeding the most atomic characters in the deep neural model, grammatical errors and misspelling keywords can be interpreted in such a way that appropriate linguistic knowledge can be learned. The systematic comparison against the four challenging benchmarking data-sets demonstrates the high level of performance obtained through our model.

Plain CNN, however, suffers from remembering long-term dependencies over sequences, as opposed to the traditional n-gram language modelling [9]. Nevertheless, according to Brown et al. [2], the system memory takes to store trigram phrases is enormous, which demands high spatial and time complexity. One possible solution to this is RNN [6], [10]. The embedded RNN hidden state and the non-linear activation function are capable of capturing long textual sequence probability distribution over iteration [6], [10]. By replacing the simple element-wise sigmoid activation function with a more complex long short-term memory (LSTM), the integrity of expressing lengthy snippets can be further improved, in line with Hochreiter and Schmidhuber [11]. Cho et al. [10] proposed a gated recurrent unit (GRU) which can dynamically reserve the state information through the pipeline of input signal. Our model benefits from Char-CNN on the regularities feature-distilling and Char-RNN upon the long-term dependencies learning.

The rest of the paper is organised as follows. Section 2 provides related literature reviews. Section 3 introduces the proposed model. Section 4 presents design of the model, data-sets, data pre-processing, and training parameters configuration. Results and detailed discussion is demonstrated in section 5. We discuss the results and findings in section 6. Finally, section 7 concludes the work and presents the possible future pointers.

## 2. Related Work

Until recently, scholars are aware that convolutional deep neural network (CDNN) can be applied to many classification tasks. The prior applications of CDNN include traffic sign classification [12], robust multi-speaker speech recognition [13], visual deep reinforcement learning from crude 3D buffer information and experience replay [14], and ImageNet classification [3]. However, these approaches have limited application in redundancy detection.

The emergence of CDNN has also shed some lights on the text mining society [15]. Kim [16] developed a two-channel CNN, a small variation of Collobert et al. model [4]. Kim's model was specifically used for sentiment polarity scoring and questions classification. Zhang et al. [7] offered an empirical investigation on the application of character-level convolutional networks for binary textual classification. Further, Yin and Schutze [17] exhibited a multi-channel CNN for sentence classification. Zhang et al. [5] gave a multiple word embeddings based CNN model for sentence categorisation. Nevertheless, those frameworks were tested merely on simple classification or sentiment analysis.

Traditionally, pure RNNs were applied to fields like statistical machine translation [10], polyphonous music modelling, speech signal understanding [18], and Python programme evaluation [19]. Currently, RNNs have seen promising results on text classification [18]. RNN is usually powered by the recurrent units such as $tanh$ [18]. Advanced recurrent units such as GRU cells [10], LSTM [11], and MGU [20] are released recently. MGU is probably better than GRU in that the gating mechanism in MGU not only halves the number of units when building up the model, but also reduce the size of the model by optimising the number of parameters during network interaction [18]. Moreover, GRU cell is different from the LSTM in that a separate memory cells is not presented in the former one. However, there is no indication which approach has the best performance without further investigation [18].

Lai et al. [21] proposed a recurrent convolutional neural network (RCNN). The recurrent structure learns the specific word with its left and right context in Lai et al. [21]. In contrast to the approach taken by Lai et al. [21], inferred sentence structure in our model is obtained through the pre-trained word2vec embeddings [22] or GloVe [23] embeddings. Kim et al. [8] described a hybrid network with CNN-LSTM to capture syntactic information buried in the contextual data. Moreover, empirical surveying by Zhang et al. shows that DCNN is less effective when the data volume is less than several million [7]. Our proposed model is the hybrid of a convolutional network and recurrent structure with character level encoding. The joint framework fits well with the four benchmarking data corpuses and yielded comparable performance.

## 3. Ingredients of the Model

Our Char-CNN was tailored from the original model to make it suitable for this work. Our Char-CNN consists of a convolving layer, RELU, and max-pooling layer. The classical Char-CNN from Vosoughi et al. [6] has multi-layer structure which seems inefficient when trained with the benchmarking data. We adapted Char-RNN from Cho et al. [10] and Hochreiter and Schmidhuber [11]. Our Char-RNN model has three different variations including GRU [10], MGU [20], and LSTM [11]. Our Char-RNN is different from the tradition in that the input gate fits with the intermediate result from Char-CNN and the output gate follows modulated activation.

### 3.1. Character-Aware Convolutional Neural Network

In our model, $C$ represents the container of character quantisation and $d$ is the character embedding size. The model was developed to support the same 70 characters supported by Vosoughi et al, which are presented below [6].

abcdefghijklmnopqrstuvwxyz0123456789
-,;.!?:'"/\|_#$%^&*~'+-=<>()[]{}

Figure 1: Illustration of 70 characters supported by our model.

Suppose that a term $t \in C$ is comprised of a sequence of characters with length $k$, and then the character-level word representation of $t$ is denoted as $C^k \in \mathbb{R}^d \times k$. Each character is encoded as a binary vector $v \in \{0,1\}^d$. Because the character quantisation is composed of 70 non-space tokens, $d$ is equal to 70 in this case.

Given the above input sequence, the 1-D convolving layer discrete kernel function $f(i) \in [1, m] \to \mathbb{R}$ and the proper padding algorithm $p$, the output feature map is

$$h^k(i) = RELU(\sum_{i=1}^{m} C^k f(i \times tk + o)), \qquad (1)$$

where $o = m - k + 1$ is a free constant, $m$ depicts the output size, and RELU stands for the rectified linear unit [24].

The temporal max-pooling layer operates in the following manner. Suppose that stride is highlighted by $s$,

$$y(i) = max_{i=1}^{m} C^k f(i \times s - k), \qquad (2)$$

Notice that padding scheme is not applied at the max-pooling layer.

## 3.2. Character-Aware Recurrent Neural Network

We now describe three RNN recurrent units, namely GRU [10], MGU [20], and LSTM [11]. Compared to the traditional implementation on GRU, MGU, and LSTM, we implanted an aggregator to produce combined result from two ingredients of CRNN.

Our LSTM is composed by five parametric arguments: memory cell $m_t^j$, input gate $i_t^j$, forget gate $f_t^j$, output gate $o_t^j$, and aggregated output state $h_t^j$ for each $j$-th element. $m_t^j$ maintains the memory information of the $j$-th LSTM unit. The update of the memory cell is formulated as

$$m_t^j = f_t^j c_{t-1}^j + i_t^j \tilde{m}_t^j. \tag{3}$$

$\tilde{m}_t^j$ is defined as

$$\tilde{m}_t^j = tanh(W_m x_t + U_m h_{t-1}). \tag{4}$$

The input gate $i_t^j$ is formulated by

$$i_t^j = \phi(W_i x_t + U_i h_{t-1} + V_i m_{t-1})^j, \tag{5}$$

where $V_i$ is a diagonal matrix and $\phi$ is an activation function. The modulation of the forget gate $f_t^j$ is as follows

$$f_t^j = \phi(W_f x_t + U_f h_{t-1} + V_f m_{t-1})^j, \tag{6}$$

where $V_f$ is a diagonal matrix. The output gate $o_t^j$ is calculated as

$$o_t^j = \phi(W_o x_t + U_o h_{t-1} + V_o m_t)^j, \tag{7}$$

where $\phi$ is a nonlinear logistic function and $V_o$ is a diagonal matrix. The output of the $j$-th LSTM element $h_t^j$ is computed as

$$h_t^j = o_t^j tanh(m_t^j), \tag{8}$$

where $o_t^j$ decides the exposed amount of memory content. Due to the adaptive forget gate, memory cell, and exposure mechanism, LSTM based Char-RNN is able to determine to what extent the variable length input sequence or output result is retained, which is analogous to the long-term memory dependency capturing.

GRU cell is different from the LSTM in that the memory content exposure is not presented. GRU cell consists of a reset gate, an update gate, and an aggregated output gate. Memory information inside the GRU cell is fully exposable [18]. Besides, the location of the input gate in the LSTM unit and the corresponding reset gate in the GRU cell are different. The modulation inside a GRU cell is described as follows. GRU cell consists of a reset gate $r_t^j$, an update gate $z_t^j$, and an output $o_t^j$. The activation at timestep $t$ of the $j$-th GRU cell is processed as a linear transformation on the previous state $o_{t-1}^j$ and the current state $\tilde{o}_t^j$:

$$\tilde{o}_t^j = (1 - z_t^j) o_{t-1}^j + z_t^j \tilde{o}_t^j, \tag{9}$$

where the update gate $z_t^j$ controls the proportion of the unit being changed. The update gate is defined as

$$z_t^j = \sigma(W_z x_t + U_z o_{t-1})^j. \tag{10}$$

The candidate output gate $\tilde{o}_t^j$ is computed as

$$\tilde{o}_t^j = tanh(W x_t + U(r_t \odot o_{t-1}))^j, \tag{11}$$

where $\odot$ is an element-wise multiplication and $r_t^j$ is the reset gate. The rest gate $r_t^j$ requires the previous recurrent output unit $o_{t-1}$ and the current input sequence $x_t$ for the computation.

$$r_t^j = \phi(W_r x_t + U_r o_{t-1})^j. \tag{12}$$

Long-term memory dependencies inside the GRU cell is accomplished by the sophisticated update mechanism. Without the reset or forget gate apparatus, the GRU cell is not able to utilise the model capacity effectively. Detailed discussion is available at Chung et al. [18].

Our MGU consists of a forget gate and an aggregated hidden recurrent output. The recent MGU has shed some insights on our model. Zhou et al.'s MGU cell has around 33% less training hyper-parameters and has equivalent performance with GRU [20].

## 4. Experimental Evaluation

In this section, we present our model design, data-sets statistics, data pre-processing, and experimental settings.

### 4.1. Model Design

TABLE 1: The list of hyper-parameters with value.

| Filters ($F$) | 400 |
|---|---|
| Hidden Size ($H$) | 400 |
| Window Size ($\Lambda$) | 20 |
| Pooling Window ($P$) | 2 |
| Stride | 1 |
| Padding Algorithm | VALID |
| Learning Rate | 0.01 |
| Training Steps | 1000 |
| Optimiser | Adam [25] |

We refer to Table 1 as the global training parameters configuration. The 'VALID' padding refers to no artificial padding, in contrast with the 'SAME' padding. The initial learning rate is set to 0.01. Moreover, the global training process is optimised through the Adam optimiser [25]. The learning rate decay is defined by Adam optimiser [25].

$$\alpha_t = \text{lr}_{t-1} \times \frac{\sqrt{1 - \beta_2^t}}{1 - \beta_1^{(t-1)}}, \tag{13}$$

where $\text{lr}_{t-1}$ is the learning rate for previous training step. $\beta_1$ is the exponential decay rate for the 1st moment estimates and $\beta_2$ is for the 2nd moment estimates [25].

In Figure 2, $n$ stands for the padded input sequence length. $\Lambda$ indicates the kernel window size. $F$ represents the number of filters applied at the convolving layer and $P$ denotes the pooling window. The default stride is 1. The operation on the 1-D temporal convolving layer, followed by the RELU activation operation, leads to the high level features with frame size $n - \Lambda + 1$. A squeeze function is performed to get a reduced matrix formation with filter size $n - \Lambda$. Subsequently, two separate branches of feature computation are conducted. On one hand, the temporal max-pooling layer of Char-CNN yields an output with feature size $(n - \Lambda)/P$. In the following layer of Char-CNN, a maximum reduction operation is operated over the sentence dimension to obtain an intermediate matrix with frame size $(n - \Lambda)/P - 1$.

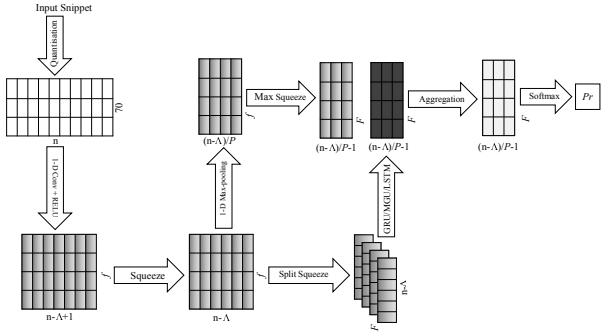

Figure 2: Illustration of our model design.

On the other hand, as the GRU/MGU/LSTM based Char-RNN only recognises input snippets in a list of sequences, we thus split and squeeze the matrix retrieved from the previous layer along with the $F$ dimension to achieve a list of corresponding sequences. A GRU/MGU/LSTM unit embedded with $H$ as the hidden size is constructed to acquire a matrix encoding. For the aggregation operation, it takes the output from the GRU/MGU/LSTM and the result matrix from the maximum reduction operation upon the Char-CNN and modulates the two encodings accordingly. Given the probability distribution $Pr$ over the possible labels of choice from the penultimate layer, the softmax will predict the target label. This completes a full training cycle for our architecture.

Towards the final output, the number of frames from the softmax layer adapts to the classification data itself. For instance, if it is a 55-class problem, then the output frame is 55. We evaluated the different conditions pair (i.e., $(0.9, 0.1), (0.8, 0.2) \ldots (0.1, 0.9)$) on the aggregation layer. We found that a large proportion assigned to the Char-CNN (i.e., 0.7) and a small distribution on the Char-RNN (i.e., 0.3) achieves the best performance. Additionally, because drop-out layer does not always appear to be effective in practice, we therefore did not include any drop-out layers in this regard [15].

### 4.2. Data and Pre-processing

As the major focus of this work is to develop CRNN for enhancing the topics classification, the appropriate benchmark data-sets would be natural textual streams or raw sentential data. We utilised four popular classification data streams from the Internet for our study. They are listed below:

- **Google-news** is a small portion of excerpts from the online platform giant Google [26], specifically the Google news channel.
- **Twenty-news-groups** was originated from the well-known American newspaper publishers formed as twenty-news-groups, probably first appeared at Lang's work [27].
- **Brown Corpus** of Standard American English, often abbreviated as the Brown Corpus [28]. The Brown Corpus encompasses with one million tokens of American English texts sampled from 15 different textual categories. The Brown Corpus is created by Francis and Kucera at Brwon University in 1960s [28].
- **Question Classification** was created by Li and Roth [29]. The question classification collection contains 50 classes of texts. The specific data-set characteristic is shown at Table 2 below.

TABLE 2: Data-sets statistics. QC stands for Question Classification collection and MST denotes mean sentence length.

| Data | Google | Twenty | Brown | QC |
|---|---|---|---|---|
| Size | 2066 | 18000 | 57340 | 5952 |
| Vocabulary | 11194 | 252999 | 55528 | 8634 |
| Total Words | 49988 | 1806249 | 649408 | 32356 |
| MST | 24.2 | 159.6 | 11.1 | 5.4 |
| Meta-data | No | Yes | Yes | Yes |
| Classes | 55 | 20 | 15 | 50 |

Table 2 reports the summary statistics based on the benchmarking data corpus. We addressed the specificity of data with a six-dimension schema:

1. the number of sentences in the data-set
2. vocabulary of the stream
3. the total number of terms in a stream
4. average sentential snippet length
5. meta-data inclusion
6. the number of labels or classes of the corpus

During pre-processing in prior input features feeding, we performed standard NLP techniques such as removing English stop words and striping off the newsgroup related meta-data, which includes noisy headers, footers, and quotes. The vocabulary processor for the Word-CNN and Word-RNN, LinearSVM [30], KNN-WMD [31], and plain KNN requires the normalised BOW features when constructing the corpus vectors. The normalised BOW generated a global uni-gram based dictionary mapping. With the presence of the uni-gram indexer, we could readily remove low frequency terms and lengthy snippets. As a general rule-of-thumb, we set the bottom frequency to 10 and the maximum length to 500. On the other hand, Char-RNN and Char-CNN employ a byte processor to map snippets into sequence of identities for bytes.

We evaluated our model and the others according to a cross-validated training/testing data split as shown in Table 3. For Google-news, the full corpus is selected. For twenty-news-groups, an arbitrary 1,100 snippets were taken from the total 18,000. 5,500 out of 57,340 were selected from the Brown Corpus and 5,500 out of 5,952 sentences were picked out from the question classification stream. Batch size defines a small amount of sentences involved in each epoch of training. For example, the batch size for the Google-news data is 50, indicating that each training epoch contains 40 samples. Similarly, the quantity of snippets for the twenty-news-groups, Brown corpus, and the question classification stream is all 20 uniformly.

TABLE 3: Experimental testing/training split. Batch size denotes the number of samples used in each training epoch.

| Data | Training | Testing | Batch Size | Total Words |
|---|---|---|---|---|
| Google | 2000 | 66 | 50 | 1032 |
| Twenty | 1000 | 100 | 50 | 655 |
| Brown | 5000 | 500 | 250 | 4568 |
| QC | 5000 | 500 | 250 | 7665 |

## 5. Results

In this paper, we constructed a novel CRNN model which employs prominent feature-filtering from Char-CNN and long-term sequence understanding from Char-RNN. Our framework automatically learns grammatical errors and misspelling through subword information. Our architecture also benefits from the latest development on GRU unit, which reduces algorithmic runtime without compromise of the performance.

In general, we showed the results independently for each benchmarking data stream. Evaluation on precision, recall, and F1 score were demonstrated for each of our data-sets. Our experiment was conducted on a cross-validated training and testing data split. The participants for our experiment are listed in Table 4. There were 20 frameworks in total for our experimental evaluation. The token random indicates the arbitrary initialisation of word vector.

As shown in Figure 3a, 3b, and 3c (Google-news), our model ranked fifth on precision rate, with 2.44% less than the best one. For the recall rate and F1 score, our model yielded first and third respectively. Our architecture achieved 2.62% more recall and 0.57% less F1 score compared with the best one.

For twenty-news-groups collection, we referred to Figure 3d, 3e, and 3f. Our algorithm achieved the optimal precision rate, recall ratio, and F1 score, leading the next best with 3.68%, 1.00%, and 2.20% respectively. Our algorithm yielded a similar precision rate and recall rate. For the F1 score, the lowest score 34.65% was from Word-RNN and random embeddings combination. Our algorithm obtained much better precision rate, recall ratio, and F1 score than Word-RNNs and Word-CNNs.

For Brown Corpus, observations can be derived from Figure 3g, 3h, and 3i. In general, our framework achieved the next best on precision rate, 0.54% less than the Char-CNN. CRNN ranked third on recall rate, having 1.40% difference to the optimum one. Our model obtained the best F1 score, with 0.05% more than the next best. From Figure 3g, 3h, and 3i, word level neural networks was worse than our model.

TABLE 4: The list of comparison algorithms and our model. The numbering process conforms with the indices appear at the x-axis of each sub-figure in Figure 3. The numbering system follows the pattern ai where a denotes algorithm and i depicts the index of that specific participant.

| Numbering | Base | Encoding |
|---|---|---|
| a1 | | GRU |
| a2 | CRNN | MGU |
| a3 | | LSTM |
| a4 | Char-CNN [7] | – |
| a5 | Char-RNN [20] | – |
| a6 | | word2vec |
| a7 | Word-CNN [15] | GloVe |
| a8 | | random |
| a9 | | word2vec |
| a10 | Word-RNN [19] | GloVe |
| a11 | | random |
| a12 | | word2vec |
| a13 | LinearSVM [30] | GloVe |
| a14 | | sent2vec [32] |
| a15 | | word2vec |
| a16 | KNN-WMD [31] | GloVe |
| a17 | | sent2vec [32] |
| a18 | | word2vec |
| a19 | KNN | GloVe |
| a20 | | sent2vec [32] |

For question classification corpus, precision rate, recall rate, and F1 score were displayed in Figure 3j, 3k, and 3l respectively. From Figure 3j, our model ranked ninth on precision rate, with 2.31% less than the best performer. For the recall ratio and F1 score, our algorithm ranked first on both, achieving 4.23% and 0.77% more than the next best algorithm.

We have previously mentioned about the performance comparison on our model over the competitors. We now analyse the performance difference on Word-CNNs, LinearSVMs [30], KNN-WMDs, and KNN series.

The average precision rate and F1 score for Word-CNN+word2vec was 0.91% and 0.32% higher than Word-CNN+GloVe. Based on such a small scale data-set, this kind of difference was significant. Similarly, for Word-RNNs, the mean precision and F1 score for word2vec based was 1.06% and 0.47% higher than the Glove sponsored. The precision rate, recall rate, and F1 score difference between Word-CNN+GloVe and Word-CNN+random was 0.31%, 0.12%, and 0.28% respectively. For Word-RNN+random and Word-RNN+GloVe, the performance difference on precision rate, recall ratio, and F1 score was 2.64%, 3.43%, and 3.10% respectively.

For non-neural networks, we emphasized the LinearSVMs, KNN-WMDs and traditional KNNs. Sent2vec [32] encoding schema tries to interpret the sentential information into a single skip-thought vector rather than word level embeddings. From the corresponding results in Figure 3, we observed that the performance of LinearSVM+sent2vec was the worst in its series for all four benchmarking collections. This phenomenon also applied to KNN-WMD+sent2vec and KNN+sent2vec.

We explain non-neural networks in the context of

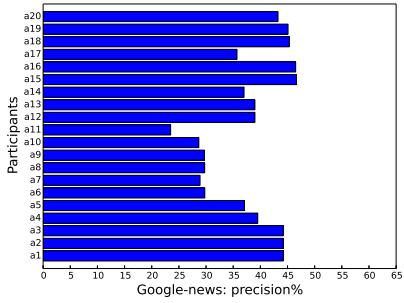
(a)
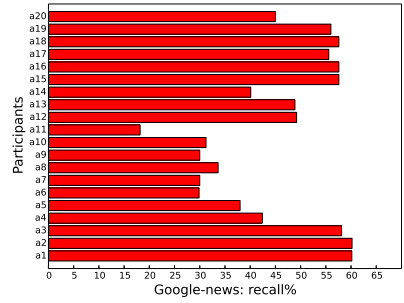
(b)
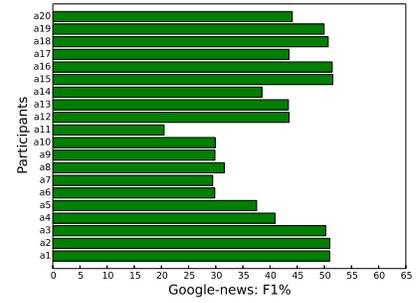
(c)
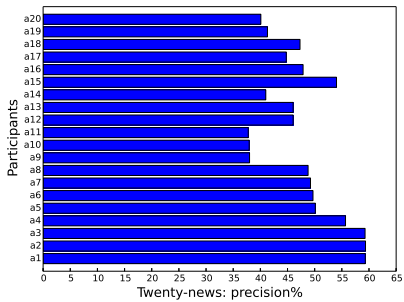
(d)
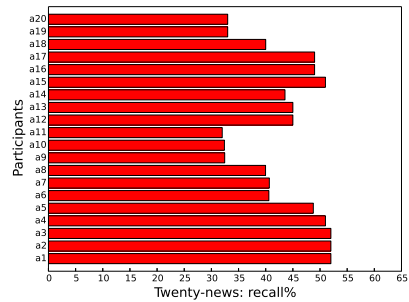
(e)
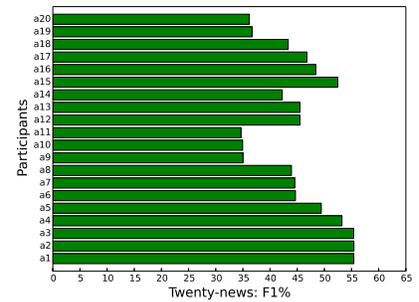
(f)
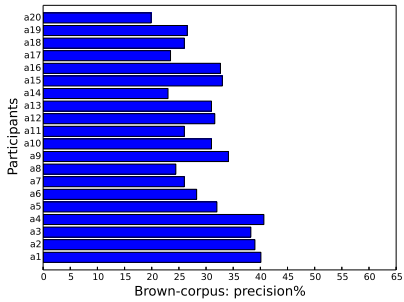
(g)
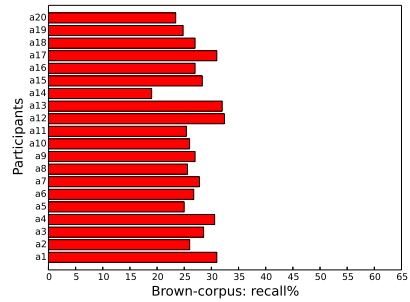
(h)
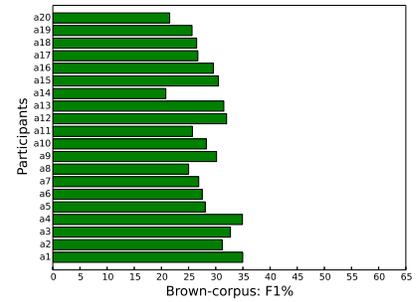
(i)
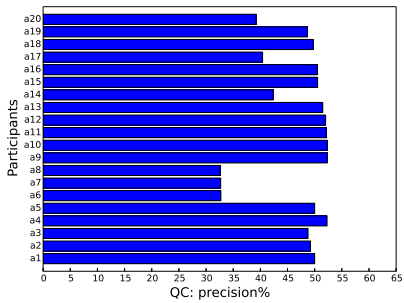
(j)
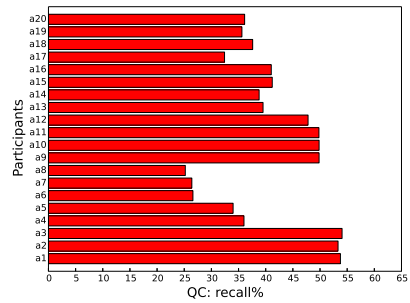
(k)
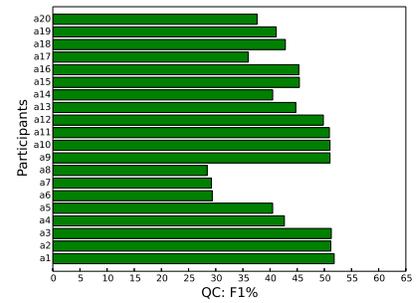
(l)

Figure 3: Precision rate, recall ratio, and F1 score statistics for the four benchmarking data-sets.

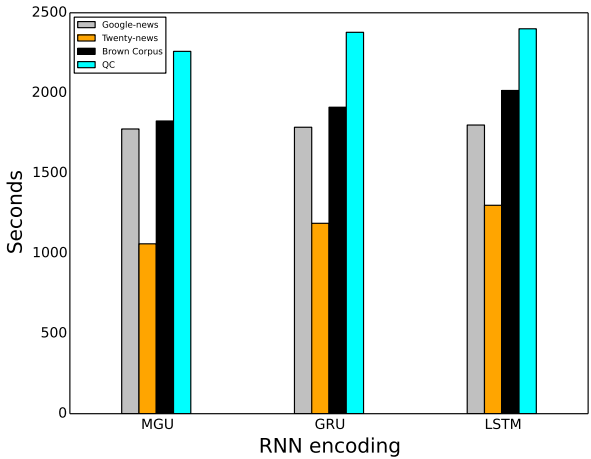

Figure 4: CRNN average runtime comparison with three different encodings.

word2vec and GloVe. For LinearSVM, both produced resemblant precision, recall, and F1 score. The only exception was question classification collection where word2vec encoding yielded 8.29% more recall and 5.11% more F1 score. For the KNN-WMDs, two embeddings posed almost identical impact under each measurement excepted the precision and F1 score on twenty-news-groups. Word2vec version yielded 6.15% more precision and 4.04% more F1 score. KNNs followed the above tendency. The only exception was twenty-news-groups of which word2vec version produced 5.99%, 7.00%, and 6.65% more precision, recall, and F1 score respectively.

We now analyse the evaluation results for KNN-WMDs and KNNs. The former one utilised WMD as the spatial distance function, the latter one applied plain cosine distance. From Figure 3, we can perceive that WMD dominant the contest over precision, recall, and F1 score. The average precision for the former one was 42.17, the latter one was 39.27. For the average recall rate, WMD sponsored models achieved 43.39% and plain KNNs obtained 37.43%. For the mean F1 score, the former one yielded 42.30%, the latter one produced 38.02%. It is obvious that WMD was much better at measuring the spatial dissimilarity.

Referring to Figure 3, LSTM, GRU, and MGU produced almost equivalent result across four data-sets. This inspired us to conduct an additional experiment involving runtime. As can be perceived from Figure 4, MGU had the minimal average runtime under each benchmarking test.

## 6. Discussion

To clarify the results in Figure 3 and 4 further, we present several findings in this section.

The most important conclusion from our experiments was that our model offered competing performance on all four data streams. For Google-news, our model obtained 2.62% more recall ratio than the next best. Our framework achieved 2.44% less precision rate and 0.57% less F1 score than the best one. For twenty-news-groups, our algorithm produced the best precision rate, recall rate, and F1 score, with 3.68%, 1.00%, and 2.20% more than the next best. For Brown Corpus, our model yielded 0.05% more F1 score than the next optimum and 0.54% less precision rate and 1.40% less recall than the optimal one. For question classification stream, our algorithm achieved the best recall rate and F1 score, having 4.23% and 0.77% more than the next best. For precision ratio, our model obtained 2.31% less than the best performer. We attributed this to the superiority of the joint two neural networks, Char-CNN and Char-RNN specifically. The aggregation layer in CRNN combined together two neural structures to improve redundancy detection. We also perceived that character-level neural models often yielded better accuracy than word-level ones. Character-level encoding can capture subword information, misspelling, and grammatical errors, which formed a rich syntactical knowledge base.

Based on the evaluation on word2vec, GloVe, and randomised word vector initialisation, we can perceive that pre-trained word vectors exerted a positive effect on the classification tasks. Further, as word2vec contains a vocabulary of three million words and the volume of GloVe is 0.4 million, we could expect that the possibility of out-of-vocabulary for word2vec is much lower than GloVe. From the evaluation results, word2vec sponsored models performed better than the others.

We emphasised the difference between sentence level vectorisation (i.e., sent2vec [32]) and word level encodings on non-neural network frameworks. We summarise that smaller embeddings unit can lead to better performance. We attributed this to the rich word-level semantics understanding. Another important finding was that the substitution of word embeddings schema posed less effect on non-neural networks than neural models. This may be due to the way sentence similarity was computed. In non-neural models, spatial distance was often determined by true spatial distance.

We also observed that KNN-WMD gained better results than KNN. We attributed this to the accurate computation of the target word to the pivot word distance and summation of the corresponding distance piles. When combining base algorithm with the pre-trained word embeddings like word2vec and GloVe, the performance of KNN-WMD and KNN can be further boosted.

Efficient runtime is important to us because CRNN involves large amount of training parameters. An effective encoding cell saves computational time and boosts throughput. From Figure 4, we observed that MUG required the minimum runtime without compromising performance. This conformed to the lowest number of training parameters within MGU.

## 7. Conclusion

In this work, we applied the latest development on the character-aware deep neural classification network, Char-

CNN and Char-RNN specifically, for the effective categorisation of redundant snippets. We extended the classical Char-CNN structure in a singleton to incorporate with the Char-RNN framework to form an efficient redundancy detection architecture. This novel framework benefits from the advantageous salient feature-filtering from the Char-CNN and the long term memory cells from the Char-RNN. We further explored the usefulness of an aggregation layer as a penultimate gate, gluing the encoding matrix generated from the Char-CNN and Char-RNN jointly for CRNN. We perceived that this enhances the detection accuracy. Evaluation on precision, recall, and F1 score indicated the efficacy of our framework. We also assessed the effects of applying different hidden units on the benchmarking data-sets and reported their performance and runtime statistics. We utilised the different word encoding schemes to deliver an exhaustive redundancy detection experiment. In the future, we hope to extend our model to a wide variety of multi-class benchmark data for a generic redundancy detection framework and performance evaluation.